\documentclass[letterpaper, 10 pt, conference]{ieeeconf}  
\IEEEoverridecommandlockouts                              

\usepackage{amsmath}
\usepackage{bbm}
\usepackage{dsfont}
\usepackage{booktabs, siunitx, multirow}

\usepackage{algorithm}

\let\labelindent\relax
\usepackage{enumitem}

\newcommand{\ours}{StratMamba}
\long\def\invis#1{}

\usepackage{xcolor}
\usepackage{amssymb}
\usepackage{graphicx}
\usepackage{mathtools}
\usepackage{amsmath}
\usepackage{gensymb}
\usepackage{float}
\usepackage{multirow}
\usepackage{makecell}
\usepackage{mathtools}
\usepackage{hyperref}
\hypersetup{
    colorlinks=true,
    linkcolor=blue,
    filecolor=magenta,      
    urlcolor=blue,
    pdftitle={Overleaf Example},
    pdfpagemode=FullScreen,
    }

\usepackage{tabularx}
    \newcolumntype{L}{>{\raggedright\arraybackslash}X}
\usepackage[utf8]{inputenc}
\usepackage[english]{babel}

\usepackage{amsthm}

\usepackage{subcaption}

\title{\LARGE \bf  \ours: Strategic and Reactive Stream Partitioning for Path-Efficient LiDAR-Based Obstacle Avoidance
}

\author{
Hung-Chieh Wu$^{1,2}$,
Xiaopan Zhang$^{2,3}$,
Kasra Sinaei$^{1,2}$,
Ryan Abnavi$^{2}$,\\
Kasun Weerakoon$^{2}$,
Christopher Bradley$^{2,4}$,
Seyed Fakoorian$^{2}$,
Jiachen Li$^{3}$,
and Donald Ebeigbe$^{1}$%
\thanks{$^{1}$The Pennsylvania State University, USA.}
\thanks{$^{2}$AlphaZ, Inc., USA.}
\thanks{$^{3}$Georgia Institute of Technology, USA.}
\thanks{$^{4}$CSAIL, Massachusetts Institute of Technology, USA.}
\thanks{This work was conducted during the authors' internship at AlphaZ, Inc. in collaboration with the AlphaZ, Inc. research team.}
}

\makeatletter
\newcommand{\copyrighttext}{%
\footnotesize
\copyright~2026 IEEE. Personal use of this material is permitted.
Permission from IEEE must be obtained for all other uses, in any current or future media,
including reprinting/republishing this material for advertising or promotional purposes,
creating new collective works, for resale or redistribution to servers or lists,
or reuse of any copyrighted component of this work in other works.
}

\def\ps@IEEEtitlepagestyle{%
  \def\@oddhead{}%
  \def\@evenhead{}%
  \def\@oddfoot{%
    \parbox{\textwidth}{\centering\copyrighttext}
  }%
  \def\@evenfoot{}%
}
\makeatother

\begin{document}

\maketitle
\thispagestyle{IEEEtitlepagestyle}
\pagestyle{empty}

\begin{abstract}
This paper proposes \ours{}, a dual-stream Mamba-based temporal modeling architecture, to more efficiently capture long-horizon temporal dependencies required for robot navigation in complex and obstacle-rich environments. \ours{} leverages a combination of fast-decay and slow-decay memory architectures, where the fast-decay component processes high-frequency LiDAR data for reactive obstacle avoidance, while the slow-decay component maintains longer-horizon goal information for strategic planning. We perform extensive evaluations of different obstacle avoidance scenarios in IsaacLab and Gazebo, while also validating successful sim-to-real deployment on a Unitree Go1 quadruped robot navigating in the presence of static/dynamic obstacles. Comparisons with other temporal RL baselines, such as LSTM, Transformer, and Vanilla-Mamba, show that our \ours{} achieves exceptional temporal reasoning efficiency with a lower timeout rate, while maintaining the fastest navigation speed (576 median steps, 5.0\% better than Vanilla-Mamba). It also achieves the highest path optimality (0.915 path efficiency) across all baselines. Real-world evaluation reveals that \ours{} maintains more robust performance across extended LiDAR ranges compared to vanilla Mamba and the Transformer, demonstrating that dual-stream partitioning effectively balances reactive safety with strategic navigation under challenging sensing conditions.
\end{abstract}

\section{Introduction} \label{sec:intro}

Mobile robots are increasingly deployed in both indoor and outdoor environments for applications ranging from logistics and delivery to inspection and exploration \cite{wang2020search-rescue,karma2015use,li2019fire}. A fundamental capability enabling autonomy in these scenarios is reliable obstacle avoidance, which ensures safety while maintaining efficient navigation toward goals. Numerous approaches have been developed to address this challenge, broadly categorized into classical methods and learning-based methods. Despite significant progress, enabling robots to exploit temporal coherence in their perception and control remains a major challenge, particularly in complex and cluttered environments.

\begin{figure}[t]
      \centering
    \includegraphics[width=\columnwidth]{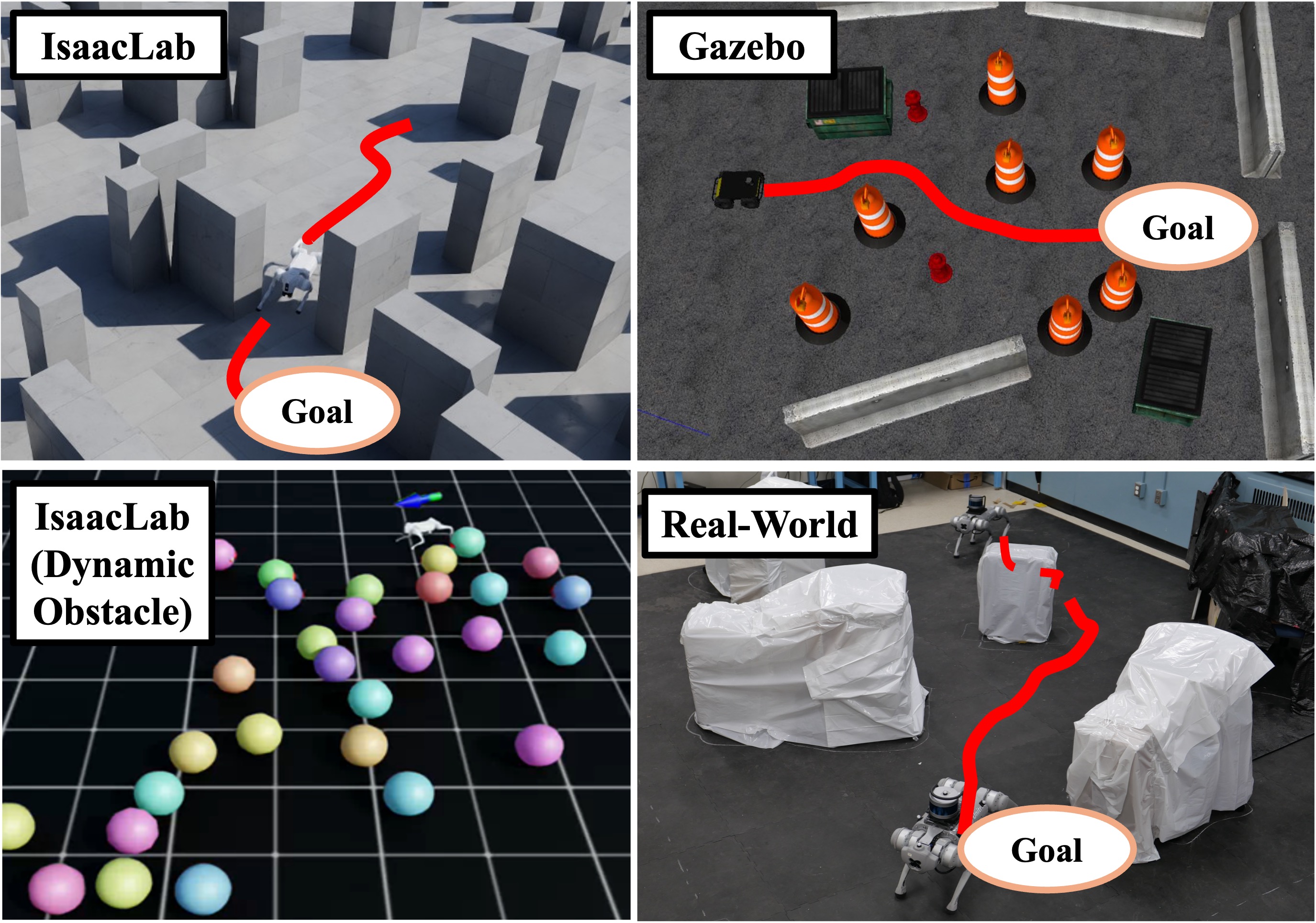}
      \caption{\small{\ours{} demonstrates robust navigation across diverse scenarios: obstacle-rich environments in IsaacLab (top-left) and Gazebo (top-right), IsaacLab dynamic obstacle avoidance (bottom-left), and Real-World deployment (bottom-right). \ours{} successfully avoids collisions with diverse obstacles, validating \ours{}'s generalization and effective sim-to-real transfer.}}
      \label{fig:cover-image}
      \vspace{-22pt}
\end{figure}

Classical methods such as Model Predictive Control (MPC) \cite{park2012mpepc,arul2024unconstrained} and Optimal Reciprocal Collision Avoidance (ORCA) \cite{orca} show success in structured settings, leveraging optimization and geometric reasoning to compute collision-free trajectories. These approaches benefit from well-defined formulations and interpretability but often struggle when faced with high-dimensional sensor data (e.g., raw LiDAR scans \cite{catapang2016obstacle}) or dynamic obstacles. Their reliance on precise modeling also limits scalability to diverse real-world scenarios, where uncertainties, sensor noise, and unmodeled dynamics are prevalent. Consequently, there has been a shift toward learning-based approaches that can directly leverage sensor data and adapt to complex environments.

Reinforcement learning (RL) has emerged as a promising alternative by learning obstacle avoidance policies through interaction with environments \cite{zhang2021reinforcement,faust2018prm,patel2021dwa,weerakoon2022terp,yao2025towards}. RL agents can generalize from raw sensor observations, enabling direct mapping from LiDAR or vision data to control commands without explicit modeling \cite{kahn2021badgr,dugas2021navrep}. However, conventional RL policies often assume independent observations at each time step, which can lead to suboptimal or unsafe behavior when the temporal context is ignored. Methods such as Long Short-Term Memory (LSTM) networks \cite{hochreiter1997long} and Transformers \cite{wolf2020transformers} have been introduced to capture temporal dependencies, but they introduce significant computational overhead, limiting deployment in real-time, resource-constrained robotic platforms.

In this work, we address these challenges by introducing \ours{}, a dual-rate state partitioning architecture that extends Vanilla-Mamba's single state-space modeling for robotic navigation. Navigation tasks inherently operate at two temporal scales: reactive obstacle avoidance requires high-frequency responses to LiDAR, while goal-directed planning demands persistent long-horizon reasoning. \ours{} structures this asymmetry by partitioning Mamba's internal state into specialized streams with complementary input routing: proximity streams process reactive information, while goal streams maintain strategic goal representations.

To isolate the contribution of the temporal encoder, we evaluate LSTM, Transformer, Mamba, and \ours{} under a common PPO framework, varying only the temporal encoder design. \ours{} achieves superior training efficiency, computational performance, and navigation robustness while retaining Mamba's linear O(L) complexity advantage over Transformer's quadratic O(L²) attention cost.

The main contributions of our work include:
\begin{itemize}
    \item We propose \ours{}, a Mamba-based RL architecture for LiDAR-based obstacle avoidance. Building on Mamba, we introduce a dual-rate state-partitioning structure that separates reactive obstacle avoidance from strategic, goal-directed planning. This design enables \ours{} to converge up to 1.5$\times$ faster than LSTM and Transformer across all sequence lengths.

    \item We demonstrate substantial performance improvements over state-of-the-art baselines: \ours{} achieves outstanding lower timeout rates (0.4\%) compared to LSTM (34.0\%), up to 13.4\% higher path efficiency, and superior trajectory smoothness with Spectral Arc Length (SPARC $\uparrow$) scores of -3.550 versus -3.624 for LSTM and -3.594 for Transformer. These improvements validate both navigation reliability and execution quality.
    
    \item We demonstrate robust cross-platform generalization from IsaacLab simulation to Gazebo sim-to-sim transfer and real-world Go1 deployment. Notably, \ours{} maintains high success rates across extended LiDAR ranges where vanilla Mamba and Transformer fail catastrophically, validating superior robustness to increased sensing complexity.
    
\end{itemize}

\section{Related Work} \label{sec:related_work}

In this section, we discuss the existing literature on obstacle avoidance for mobile robots and temporal sequence modeling techniques used in robot navigation.

\begin{figure*}[t]
\vspace{4pt}
      \centering
      \includegraphics[width=0.97\textwidth]{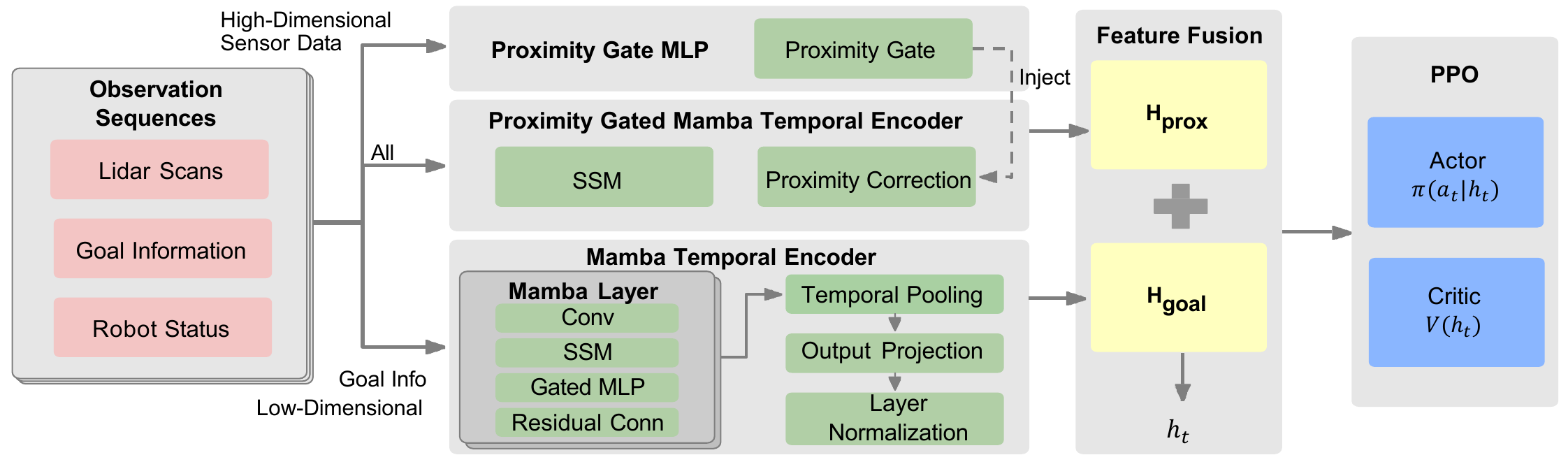}
    \caption{\small{Overview of \ours{}. Observation sequences are partitioned into a \textit{Proximity stream} for reactive obstacle avoidance and a \textit{Goal stream} for strategic navigation. Their encoded representations, $\mathbf{H}_{\text{prox}}$ and $\mathbf{H}_{\text{goal}}$ , are fused into $h_t$ and fed to PPO's actor-critic networks for action generation and value estimation. This dual-rate design enforces temporal specialization, enabling coordinated fast-reactive and slow-strategic reasoning.}}
      \label{fig:System_arch}
      \vspace{-20pt}
\end{figure*} 

\subsection{Obstacle Avoidance Methods}
Classical approaches to obstacle avoidance have provided a strong foundation for robotic navigation. Methods such as MPC optimize a sequence of control actions over a finite horizon, allowing the agent to anticipate and react to future states using a predictive model \cite{park2012mpepc,arul2024unconstrained}. Geometric techniques like ORCA enable multi-agent collision avoidance by cooperatively selecting optimal velocities in a decentralized manner \cite{van2011reciprocal, alonso2013optimal, guo2021vr}. However, these methods often rely on simplified world models and predefined dynamics, limiting their applicability. They typically struggle with high-dimensional, unstructured sensor data, such as raw LiDAR or camera inputs, and may fail in complex, cluttered real-world environments with unforeseen dynamics \cite{catapang2016obstacle,iqbal2020simulation,bijelic2020seeing}.

To overcome these limitations, RL has emerged as a powerful, data-driven alternative. By framing navigation as a sequential decision-making problem, RL enables the development of control policies that map raw sensory input directly to actions through end-to-end learning \cite{zhang2021reinforcement,faust2018prm,patel2021dwa,weerakoon2022terp,kahn2021badgr}. An agent learns optimal behavior through trial and error, guided by a reward signal, without requiring an explicit model of the world's physics. This paradigm shift allows policies to learn complex, non-linear relationships and adapt to the nuances of real-world sensor data. Our work builds upon this RL framework, specifically investigating the architectural improvements required to make RL policies not only effective, but also computationally efficient at execution time and capable of coherent, long-horizon reasoning.

\subsection{Temporal Sequence Modeling in Navigation}
Effective navigation in dynamic environments is fundamentally a problem of temporal reasoning. An agent must not only perceive its immediate surroundings but also contextualize this information within a history of observations to infer the velocity of moving objects, anticipate trajectories, and remember the layout of areas that are temporarily occluded. Standard architectures, such as Multi-Layer Perceptrons (MLPs), lack the capacity for such reasoning.

Early work addressed this by incorporating recurrent architectures, such as LSTMs, which maintain a hidden state updated at each time step, allowing integration of information over time \cite{hochreiter1997long, song2023improved, wang2025enhancing}. More recently, Transformers have gained prominence due to their self-attention mechanism, which allows every point in an observation history to attend to every other point \cite{chen2021decisiontransformer}. This facilitates the capture of complex, long-range temporal dependencies often missed by recurrent models. However, this core mechanism is also a key limitation in real-time robotics: computational and memory costs scale quadratically with sequence length ($O(L^2)$), creating a significant bottleneck in low-latency applications.

To address this efficiency challenge, recent research has turned to State Space Models (SSMs). Architectures such as Mamba build upon SSMs by incorporating a selection mechanism that allows the model to selectively focus on the most relevant information, achieving performance competitive with Transformers while maintaining linear time complexity ($O(L)$) \cite{gu2023mamba}. By efficiently compressing long historical sequences into a compact latent state, these models enable long-horizon reasoning without incurring prohibitive computational overhead, thereby making them well-suited for real-time decision-making and control tasks in robotics.

\subsection{Mamba-based Methods}
Following the introduction of the selective state-space model~\cite{gu2023mamba}, Mamba-based architectures have rapidly emerged as efficient alternatives to Transformers across diverse application domains. In computer vision, Vision Mamba~\cite{zhu2024vision} demonstrated that bidirectional SSM can achieve competitive performance in object detection and semantic segmentation while offering significantly improved computational efficiency—up to 2.8$\times$ faster than Transformers. This success has inspired numerous domain-specific adaptations, including Spatial-Mamba~\cite{xiao2024spatial}, PixMamba~\cite{lin2024pixmamba}, and PyramidMamba~\cite{wang2025pyramidmamba} for different application domains. 

Beyond single-modality applications, researchers have developed specialized multi-stream architectures to handle heterogeneous data. Fusion-Mamba~\cite{xie2024fusionmamba} proposed dual-stream feature extraction with gated attention-based fusion for RGB-infrared object detection. The multi-stream architectures reveal a key insight: while vanilla Mamba excels at temporal sequence modeling, complex tasks often benefit from architectural specialization, with different processing streams handling distinct types of information. In robotics, Mamba-based approaches remain largely unexplored for robot navigation tasks, where real-time sensor processing and obstacle avoidance are critical.

Moreover, existing robotic applications primarily employ single-stream Mamba encoders that process all input modalities uniformly, without explicit architectural mechanisms to separate fast-changing reactive information from slow-changing strategic information. As dual-stream mamba has demonstrated effectiveness in vision and multimodal domains, it is rarely implemented in real-world robotic systems, particularly for online navigation tasks that require coordinated reactive and deliberative reasoning. Motivated by these gaps and the success of specialized multi-stream architectures, we propose \ours{}, which extends Mamba to robot navigation through Dual-Rate State Partitioning. 
\section{Background: Mamba-Based Network}\label{sec:background}
In this section, we review the standard single-stream Mamba architecture; Section \ref{subsec:our_architecture} presents our dual-stream extension that enables specialized processing of reactive and strategic information. Mamba \cite{gu2023mamba} uses a state-space model (SSM) architecture designed for efficient sequence modeling. Unlike recurrent networks such as LSTMs or attention-based models like Transformers, Mamba leverages linear state-space dynamics augmented with input-dependent gating to capture long-term temporal dependencies while maintaining linear computational complexity in sequence length.

Its core is a continuous-time linear SSM: \vspace{-3pt}
\begin{equation}
\frac{d}{dt} x(t) = A x(t) + B u(t), 
\qquad 
y(t) = C x(t),
\end{equation}
where \(x(t) \in \mathbb{R}^d\) denotes the hidden state, \(u(t)\) the input, and \(y(t)\) the output. The matrices \(A, B, C\) parameterize the system, capturing the dynamics, input, and output relationships. Discretizing this system with a step size \(\Delta\) yields
\begin{equation}
x_k = \bar{A} x_{k-1} + \bar{B} u_k, 
\qquad 
y_k = C x_k,
\end{equation}
where the discrete-time transition and input matrices are \(\bar{A} = \exp(A \Delta)\) and \(\bar{B} = (\exp(A \Delta)-I)A^{-1}B\), and \(x_k, u_k, y_k\) denote the state, input, and output at step \(k\).

A key innovation in Mamba is the introduction of \emph{input-dependent gating} within the state-space layer. Each observation \(o_k\) is first embedded as \(z_k = E(o_k) \in \mathbb{R}^{d_{\text{model}}}\), and the Mamba layer updates the hidden state according to,
\begin{equation}
x_k = \bar{A}(z_k) x_{k-1} + \bar{B}(z_k) z_k,
\qquad
h_k = C(z_k) x_k,
\end{equation}
where \(\bar{A}(\cdot), \bar{B}(\cdot), C(\cdot)\) are input-dependent transformations, often implemented as linear projections modulated by gating functions. The output at step \(k\) is \(h_k\). Multiple Mamba layers are stacked, with the final hidden state used for downstream policy learning. For temporal modeling, the embedding sequence is denoted by \(Z = [z_{t-n_{\text{seq}}+1}, \dots, z_t]\).

For comparison, standard Transformers operate based on the self-attention mechanism: \vspace{-7pt}
\begin{equation}
\mathrm{Attn}(Q,K,V) = \mathrm{softmax}\!\left(\frac{QK^\top}{\sqrt{d_k}}\right) V,
\end{equation}
where \(Q, K, V\) are the query, key, and value matrices, and \(d_k\) is the dimension of the key vectors. This operation scales quadratically with sequence length due to pairwise interactions, making it computationally expensive for long sequences such as LiDAR scan histories. In contrast, Mamba processes sequences with \(\mathcal{O}(n)\) complexity, thanks to the recurrence in its state-space formulation.

In summary, given the embedding sequence \(Z\), stacked Mamba layers apply the recurrence above across time steps to capture long-range temporal dependencies. Compared to LSTMs, Mamba retains memory over longer horizons more effectively, and compared to Transformers, it is significantly more efficient in time and memory, making it well-suited for real-time robotic control.
\section{ \ours: Strategic and Reactive Stream\\
Mamba Policy Optimization } \label{sec:our-method}

In this section, we outline the methodology of \ours{}, illustrated in Fig.~\ref{fig:System_arch}. The proposed framework consists of five key components: observation space design, policy network architecture, action distribution, reward formulation, and policy training. The action distribution governs how actions are sampled and executed, while the reward formulation encodes the learning objectives into quantitative feedback. Finally, the policy training procedure is used to obtain the learned navigation policy. Each component is described in detail below.

\subsection{Observation Space}
At each discrete time step $t$, the agent receives an observation vector $o_t \in \mathbb{R}^{d}$ encoding geometric and dynamic information. The observation includes raw LiDAR ranges $r_t^{(i)}$ normalized by maximum sensor range $r_{\max}$, goal position relative to the robot's body frame, normalized distance to goal (where $p_g^w$ and $p_0^w$ denote goal and initial positions in world frame), and base velocity normalized by platform-specific limit $v_{\max}$. 
Table \ref{tab:observation_space} summarizes these components, processed as a temporal sequence $O_t = [o_{t-n_{\text{seq}}+1}, \ldots, o_t]$.

\begin{table}[t]
\vspace{3pt}
\centering
\caption{\small{Observation Space Components}}
\label{tab:observation_space}
\renewcommand{\arraystretch}{1.5}
\small
\begin{tabular}{m{2.25cm}|m{0.75cm}|m{4.2cm}}
\hline
\textbf{Component} & \textbf{Dim.} & \textbf{Definition} \\
\hline
LiDAR Scan & $\mathbb{R}^{360}$ & $l_t = [\tilde{r}_t^{(1)}, \ldots, \tilde{r}_t^{(360)}]$ where $\tilde{r}_t^{(i)} = \min(r_t^{(i)}, r_{\max})/r_{\max}$ \\
\hline
Goal Position & $\mathbb{R}^{2}$ & $g_t = (x_t^g, y_t^g)$ in robot's local body frame \\
\hline
Distance to Goal & $\mathbb{R}$ & $d_t = \|g_t\|_2 / \max(\|p_g^w - p_0^w\|_2, \epsilon)$ where $\epsilon = 10^{-5}$ \\
\hline
Base Velocity & $\mathbb{R}^{3}$ & $\tilde{v}_t^b = v_t^b/v_{\max} \in [-1,1]^3$ \\
\hline
\hline
\multicolumn{3}{c}{\textbf{Whole Observation:} $o_t = [\tilde{v}_t^b, g_t, d_t, l_t] \in \mathbb{R}^{366}$} \\
\hline
\end{tabular}
\vspace{-15pt}
\end{table}

\subsection{Dual-Stream Policy Architecture} \label{subsec:our_architecture}

Robotic navigation involves observations with fundamentally different temporal characteristics. High-dimensional LiDAR scans ($l_t \in \mathbb{R}^{360}$) change rapidly at each timestep, requiring reactive processing for obstacle avoidance, while low-dimensional goal information ($[g_t, d_t] \in \mathbb{R}^{3}$) remains relatively stable, guiding strategic planning over longer horizons. Standard temporal encoders process all components uniformly, failing to exploit this inherent structure.

We introduce \ours{}, as shown in Figure~\ref{fig:System_arch}, which organizes the observation vector $o_t = [\tilde{v}^b_t, g_t, d_t, l_t]$ into two complementary streams with asymmetric input routing:

\begin{itemize}
    \item \textbf{Proximity stream (reactive):} $o^{\text{prox}}_t = [\tilde{v}^b_t, g_t, d_t, l_t] \in \mathbb{R}^{366}$ processes all observations through a Mamba encoder where obstacle proximity adaptively modulates SSM temporal processing.
    \item \textbf{Goal stream (strategic):} $o^{\text{goal}}_t = [g_t, d_t] \in \mathbb{R}^{3}$ processes only goal information for long-horizon planning.
\end{itemize}

Each stream processes its observation history through an independent Mamba-based SSM. For the proximity stream:
\begin{equation}
Z^{\text{prox}} = [z_{t-n_{\text{seq}}+1}^{\text{prox}}, \dots, z_t^{\text{prox}}], \quad z_i^{\text{prox}} = E_{\text{prox}}(o_i^{\text{prox}}) \in \mathbb{R}^{d_{\text{model}}}
\end{equation}

where $E_{\text{prox}}: \mathbb{R}^{366} \to \mathbb{R}^{256}$ is a learned linear embedding. To enable obstacle-aware modulation, we first compute proximity gate signals $\Gamma = [\gamma_{t-n_{\text{seq}}+1}, \dots, \gamma_t]$, where $\gamma_i = \text{MLP}_{\text{gate}}(l_i) \in \mathbb{R}^{d_{\text{state}}}$ maps each LiDAR scan $l_i \in \mathbb{R}^{360}$ to a gating signal via a compact 2-layer MLP (360 $\to$ 128 $\to$ $d_{\text{state}}$). During SSM processing in each Mamba layer, the gate signal $\gamma_i$ modulates the hidden state at timestep $i$: $x_{\text{ssm}}^{(i)} \leftarrow x_{\text{ssm}}^{(i)} + W_{\text{gate}} \tanh(\gamma_i)$, where $W_{\text{gate}}$ is zero-initialized to ensure \ours{} equals vanilla Mamba at initialization. The gated sequence is processed by $K=2$ stacked Mamba layers: \vspace{-8pt}

\begin{equation}
H^{\text{prox}}
=
\text{Mamba}^{\text{prox}}_K
(
\cdots
\text{Mamba}^{\text{prox}}_1
(
Z^{\text{prox}};\Gamma
)
)
\end{equation}
When obstacles are close, the learned gate adds corrective signals for reactive processing; in open space, corrections remain near zero, preserving standard SSM behavior.

The goal stream follows an analogous formulation with independent parameters:
\begin{align}
Z^{\text{goal}} &= [E_{\text{goal}}(o_i^{\text{goal}})]_{i=t-n_{\text{seq}}+1}^{t}, \quad E_{\text{goal}}: \mathbb{R}^{3} \to \mathbb{R}^{64} \\
H_{\text{goal}} &= \text{Mamba}^{\text{goal}}_K(\cdots \text{Mamba}^{\text{goal}}_1(Z^{\text{goal}}))
\end{align}
Each Mamba layer integrates causal convolution for short-term context, compact SSM for long-term memory, gated MLP for nonlinearity, and residual connections for stable training, as illustrated in Figure \ref{fig:System_arch}.

The streams are fused additively with a zero-initialized projection: \vspace{-12pt}
\begin{equation} 
h_t = h_t^{\text{prox}} + W_{\text{fuse}} h_t^{\text{goal}} \in \mathbb{R}^{d_{\text{model}}}
\end{equation}
where $h_t^{\text{prox}} = H_{\text{prox}}[-1]$,
$h_t^{\text{goal}} = H_{\text{goal}}[-1]$,
and $W_{\text{fuse}}$ is zero-initialized. This representation $h_t$ then feeds into the actor and critic networks for action generation and value estimation, maintaining compatibility with PPO training.

\subsection{Action Distribution}
The policy head maps the hidden state \( h_t \) into the parameters of a Gaussian distribution over continuous actions: \vspace{-2pt}
\begin{equation}
a_t = (v_t, \omega_t) \sim \pi_\theta(a_t \mid O_t) = 
\mathcal{N}\!\big(\mu_\theta(h_t), \, \Sigma_\theta(h_t)\big)
\end{equation}
where \( \mu_\theta(h_t) \in \mathbb{R}^2 \) represents the mean linear velocity \( v_t \) and angular velocity \( \omega_t \), while \( \Sigma_\theta(h_t) = \mathrm{diag}(\sigma_\theta^2(h_t)) \) is the diagonal covariance matrix.

\begin{figure*}[ht!]
    \vspace{4pt}
      \centering
      \includegraphics[width=0.97\textwidth]{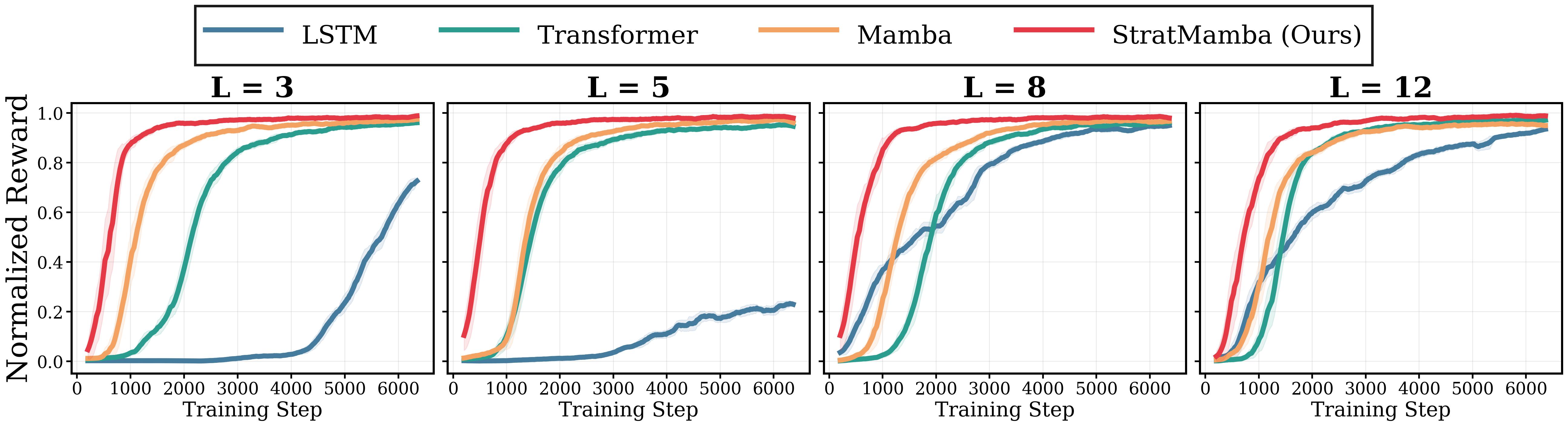}
        \caption{\small{Normalized training reward of LSTM, Transformer, Mamba, and \ours{} across sequence lengths $L \in \{3, 5, 8, 12\}$. \ours{} consistently achieves the fastest convergence and reaches near-optimal performance across horizons.}}
      \label{fig:reward_backbone_comparison}
      \vspace{-12pt}
\end{figure*}

\subsection{Reward Functions}
The training process uses continuous rewards for step-wise guidance and terminal rewards for episode outcomes. Continuous rewards include position tracking (encouraging progress toward goal $g_t$), obstacle avoidance penalty (based on closest LiDAR distance $d_t^{(i)}$), and heading alignment (measuring angular deviation $\theta_t$ from goal direction). Terminal rewards provide sparse signals for goal reaching (threshold $\delta_g$), collision detection, and premature stopping. Table \ref{tab:reward_functions} summarizes these components, combined as: $r_t = w_{\text{pos}} r^{\text{pos}}_t + w_{\text{obs}} r^{\text{obs}}_t + w_{\text{head}} r^{\text{head}}_t + w_{\text{goal}} r^{\text{goal}} + w_{\text{coll}} r^{\text{coll}} + w_{\text{stop}} r^{\text{stop}}$
\subsection{Training Details}
For large-scale policy training, we utilize the IsaacLab simulator, which generates procedurally obstacle-rich environments. Each environment spans a $20 \times 20$~m workspace populated with discrete, box-shaped obstacles whose sizes and placements are randomized to ensure diversity across training runs. We train a total of 256 simulated Unitree~Go2 robots in parallel, facilitating rapid data collection and stable policy optimization. All training is conducted on NVIDIA GPUs, with a navigation control frequency of 5~Hz and a locomotion control frequency of 50~Hz.

\begin{table}[t]
\centering
\caption{\small{Reward Function Components}}
\label{tab:reward_functions}
\renewcommand{\arraystretch}{1.5}
\small
\begin{tabular}{m{2.5cm}|m{5.3cm}}
\hline
\textbf{Component} & \textbf{Formula} \\
\hline
\multicolumn{2}{c}{\textit{Continuous Rewards (at each timestep)}} \\
\hline
Position Tracking & $r^{\text{pos}}_t = 1 - \tanh(\|g_t\|_2/\sigma)$ where $\sigma=1.0$ \\
\hline
Obstacle Penalty & $r^{\text{obs}}_t = -\max_i \frac{1}{\max(d_t^{(i)}, \epsilon)}$ where $\epsilon=0.1$ \\
\hline
Heading Penalty & $r^{\text{head}}_t = -\tanh(|\theta_t|/\sigma_\theta)$ 
\newline where $\theta_t = \arctan2(y_t^g, x_t^g)$, $\sigma_\theta=0.5$ \\
\hline
\multicolumn{2}{c}{\textit{Terminal Rewards (at episode end)}} \\
\hline
Goal Reached & $r^{\text{goal}} = \mathbf{1}\{\|g_t\|_2 < \delta_g\}$ where $\delta_g=0.5$ \\
\hline
Collision & $r^{\text{coll}} = \mathbf{1}\{\text{base contact}\}$ \\
\hline
Stop Early (Stuck) & $r^{\text{stop}} = \mathbf{1}\{\text{stopped before goal}\}$ \\
\hline
\end{tabular}
\vspace{-18pt}
\end{table}

\vspace{-3pt}

\subsection{Backbone Architecture Comparison}

To isolate the contribution of temporal modeling, we replace only the temporal sequence encoder within an otherwise identical PPO framework, comparing LSTM, Transformer, Mamba, and our proposed \ours{}. Figure~\ref{fig:reward_backbone_comparison} reports normalized training reward across sequence lengths $L \in \{3,5,8,12\}$. \ours{} generally achieves faster convergence, particularly at shorter horizons.

These results highlight the importance of structured temporal modeling. Increasing the observation horizon enriches state information by capturing obstacle approach dynamics, motion trends, and corridor geometry in partially observable settings. However, effectively exploiting this richer context requires stable long-range credit assignment. Standard recurrence degrades gradually with increasing horizon, and attention-based models become harder to optimize as sequence length grows. In contrast, the structured SSM formulation of Mamba enables linear $\mathcal{O}(L)$ temporal processing, while the proposed dual-rate decomposition in \ours{} further separates reactive and strategic dynamics.

\section{Simulation Results and Analysis}

\subsection{Simulation Environments} \label{subsuc:Testing_Scenarios}
To evaluate the navigation performance of trained policies, we deploy the policies in distinct simulators:

\begin{enumerate}
    \item \textbf{IsaacLab:}  
    All trained policies are evaluated in IsaacLab under a systematic set of dynamic obstacle scenarios, as shown in Figure~\ref{fig:Scenarios_1}.  Each episode places the robot at the origin $(0, 0)$ with a fixed goal at $(8, 8)$ in the local frame, requiring the robot to navigate an $8\,\mathrm{m}$ displacement in both axes. Each episode is limited to 15 seconds of completion time; episodes exceeding this budget are recorded as timeouts.

    \begin{figure}[t!]
          \centering
    \includegraphics[width=0.97\columnwidth]{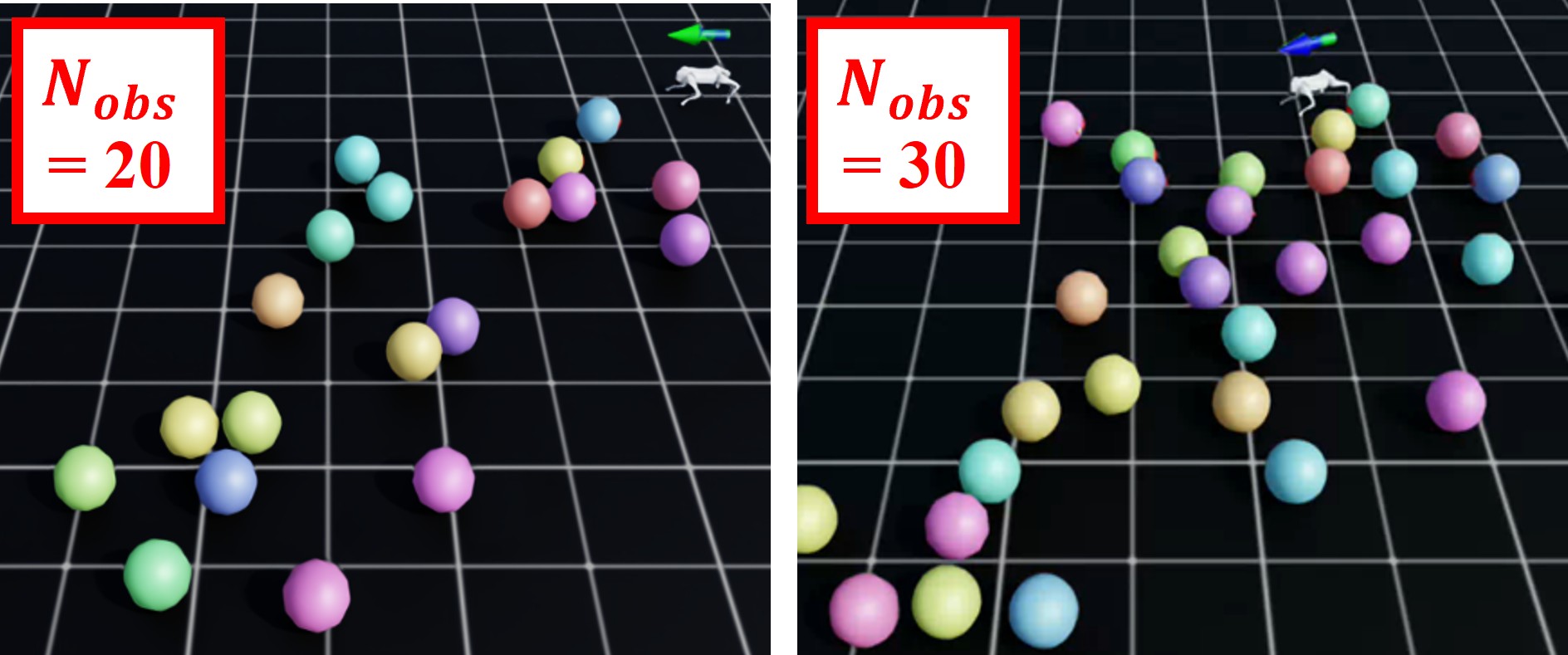}
          \caption {\small{IsaacLab evaluation with Unitree Go2 in dynamic obstacle environments. The two scenarios illustrate environments with varying dynamic obstacle densities.}}
          \label{fig:Scenarios_1}
          \vspace{-10pt}
    \end{figure} 

    \begin{figure}[t!]
          \centering
          \includegraphics[width=0.97\columnwidth]{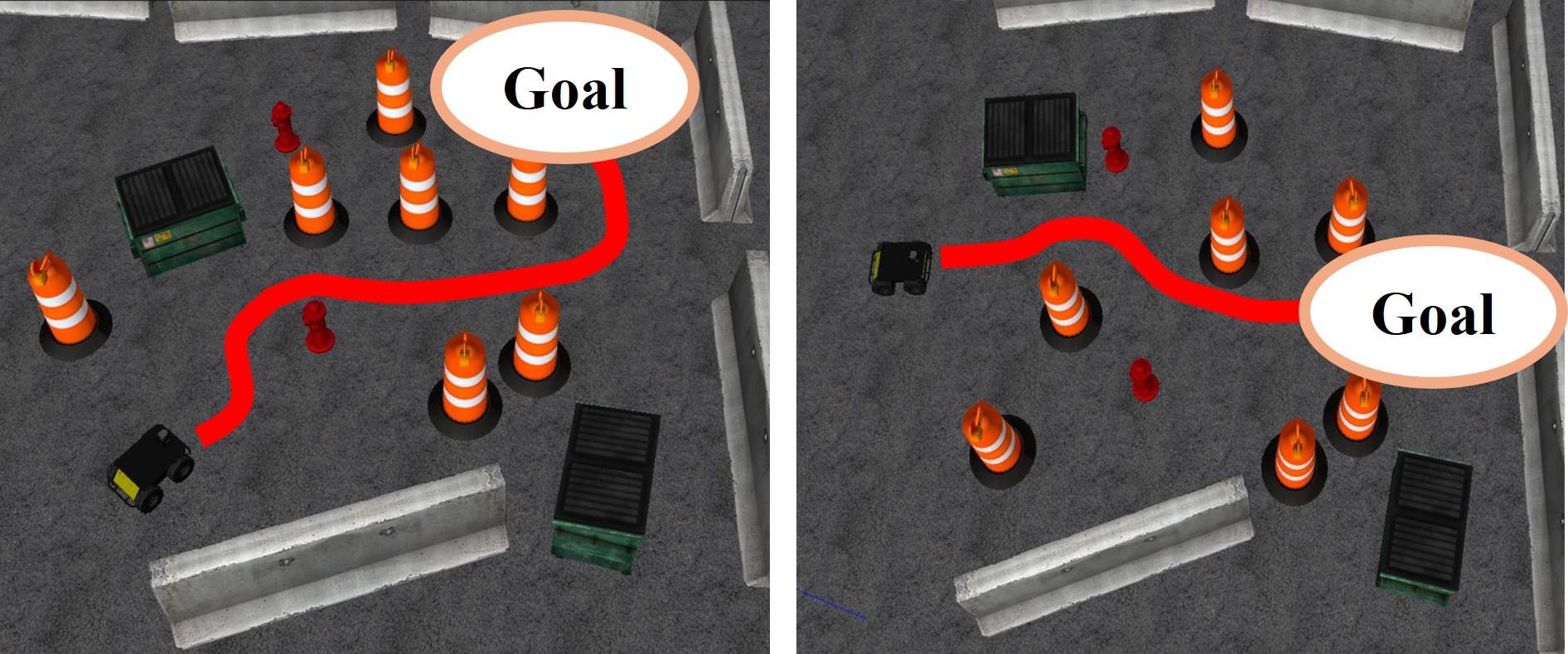}
          \caption {\small{Gazebo evaluation with the Clearpath Husky Wheeled Platform in different obstacle layouts.}}
          \label{fig:Scenarios_2}
          \vspace{-18pt}
    \end{figure}
    
    \item \textbf{Gazebo:}  
    To further evaluate cross-domain generalization, we deploy the trained policies in the Gazebo simulator via a Clearpath Husky wheeled robot. The test environments feature irregular layouts and obstacle fields distinct from the IsaacLab training maps. As shown in Figure~\ref{fig:Scenarios_2}, the Husky successfully navigates through different obstacle-rich layouts, demonstrating the policies' adaptability beyond their training domain.
    \end{enumerate}

\subsection{Evaluation Metrics} \label{subsec:Evaluation_Metrics}
To quantify the performance, several evaluation metrics are defined and organized into three perspectives:
\begin{enumerate}[leftmargin=*]
    \item \textbf{Basic Competence} evaluates fundamental navigation reliability:
    \begin{itemize}
        \item \textbf{Success Rate} ($\uparrow$): Fraction of episodes reaching the goal within the time budget without collision.
        \item \textbf{Timeout Rate} ($\downarrow$): Fraction of episodes exceeding the time budget, indicating failure to find feasible paths within temporal constraints.
    \end{itemize}

    \item \textbf{Navigation Quality} assesses path efficiency for successful episodes:
    \begin{itemize}[leftmargin=*, nosep]
        \item \textbf{Median Steps to Goal} ($\downarrow$): Representative time 
        required to reach the goal.
        \item \textbf{Path Efficiency} ($\uparrow$): Ratio of the straight-line distance from start to goal over the actual path length traveled, 
        evaluated on successful episodes only:
        \begin{equation}
        \text{PE}_i = \frac{\|\mathbf{p}^{\text{goal}} - \mathbf{p}^{\text{start}}\|_2}{P_i}
        \label{eq:path_efficiency}
        \end{equation}
        where $P_i$ is the total path length executed by the robot in episode $i$.
    \end{itemize}
    
    \item \textbf{Trajectory Smoothness} quantifies motion quality via velocity and jerk metrics. Let $\mathbf{v}(t) \in \mathbb{R}^2$ denote planar velocity, $\mathbf{x}(t)$ position, and $P = \int_{0}^{T} \lVert \mathbf{v}(t) \rVert \, dt$ the total path length.
    \begin{itemize}[leftmargin=*]
        \item \textbf{LDLJ} ($\uparrow$): Log-Dimensionless Jerk quantifies acceleration smoothness through jerk $\mathbf{j}(t) = d^{3}\mathbf{x}(t)/dt^{3}$ in a scale-invariant form: \vspace{-10pt}
        
        \begin{equation} 
        \mathrm{LDLJ}
        =
        -\log_{10}
        \left(
        \frac{T^5}{P^2}
        \int_0^T
        \|\mathbf{j}(t)\|_2^2\,dt
        \right)
        \end{equation}

        Less negative values indicate smoother motion with reduced acceleration changes.
        
        \item \textbf{SPARC} ($\uparrow$): Spectral Arc Length~\cite{beck2018sparc} measures velocity smoothness in the frequency domain. Given normalized speed $\tilde{s}(t) = s(t)/\max_t s(t)$ where $s(t) = \lVert \mathbf{v}(t)\rVert$, and spectrum magnitude $M(\omega)$ normalized to $\Omega, M \in [0,1]$ with cutoff $\Omega_c$ where $M(\Omega_c) = \tau$:
        \begin{equation}
        \mathrm{SPARC} = -\int_{0}^{\Omega_c} 
        \sqrt{1 + \left(\frac{dM}{d\Omega}\right)^2} \, d\Omega
        \label{eq:sparc}
        \end{equation}
        Values closer to zero indicate smoother trajectories with less high-frequency content.
    \end{itemize}
\end{enumerate}

\subsection{Evaluation Result} \label{subsuc:simulation_evaluation_Result}
\textbf{Basic Competence:} As shown in Table~\ref{tab:basic_competence}, \ours{} achieves the lowest timeout rate (0.4\%) while maintaining competitive success across sequence lengths, with Mamba-based methods substantially outperforming LSTM (67.3\% vs. 39.1\% average success). On the other hand, the timeout metric indicates navigation reliability: \ours{}'s consistently low rates suggest effective long-horizon planning, whereas LSTM's 34.0\% timeout reveals difficulty maintaining coherent strategies over extended temporal horizons. 

\begin{table}[t!]
    \vspace{3pt}
    \centering
    \caption{\small{Basic navigation competence evaluation across sequence lengths. \textbf{Success Rate}~$\uparrow$ measures overall goal-reaching performance. \textbf{Timeout Rate}~$\downarrow$ indicates failure to find feasible paths within the episode budget. Best in \textbf{bold}, second-best \underline{underlined}.
    }}
    \label{tab:basic_competence}
    \Huge
    \renewcommand{\arraystretch}{1.3}
    \resizebox{\columnwidth}{!}{
    \begin{tabular}{
    l
    cccc c
    cccc c
    }
    \toprule
    \multirow{2}{*}{\textbf{Method}}
    & \multicolumn{5}{c}{\textbf{Success Rate (\%)} ($\uparrow$)}
    & \multicolumn{5}{c}{\textbf{Timeout Rate (\%)} ($\downarrow$)} \\
    \cmidrule(lr){2-6} \cmidrule(lr){7-11}
    & $L$=3 & $L$=5 & $L$=8 & $L$=12 & \textbf{Avg}
    & $L$=3 & $L$=5 & $L$=8 & $L$=12 & \textbf{Avg} \\
    \midrule
    LSTM
    & 0.0 & 25.3 & 65.5 & \underline{65.5} & 39.1
    & --- & 48.2 & \underline{0.7} & 1.7 & 34.0 \\
    Transformer
    & 60.3 & 63.1 & 58.9 & 63.0 & 61.3
    & 1.4 & 1.0 & 1.0 & \underline{0.9} & 1.1 \\
    Mamba
    & \textbf{70.8} & \underline{69.6} & \textbf{72.2} & \textbf{75.9} & \textbf{72.1}
    & \underline{0.4} & \underline{0.7} & \textbf{0.5} & 1.5 & \underline{0.8} \\
    \textbf{\ours{}}
    & \underline{67.6} & \textbf{69.9} & \underline{68.9} & 62.9 & \underline{67.3}
    & \textbf{0.3} & \textbf{0.2} &  \underline{0.7} & \textbf{0.2} & \textbf{0.4} \\
    \bottomrule
    \end{tabular}
    }
    \vspace{-7pt}
    \end{table}

\textbf{Navigation Quality:}  
With respect to path quality, \ours{} achieves the fastest navigation, requiring 576 median steps, and the highest path efficiency of 0.915, as shown in Table~\ref{tab:navigation_quality}. Compared with Mamba, this represents 5.0\% fewer steps and 5.8\% higher path efficiency, with larger gains over Transformer and LSTM. At the longest horizon $L{=}12$, \ours{} reaches 0.932 path efficiency, approaching theoretical optimality. These results demonstrate that \ours{}'s dual-stream architecture excels at long-horizon trajectory optimization, producing not only faster but also more direct paths that reduce mission time in deployment scenarios.

\begin{table}[t!]
    \centering
    \caption{\small{
    Navigation efficiency evaluation on successful episodes across sequence lengths. \textbf{Median Steps to Goal}~$\downarrow$ measures the time to reach the goal. \textbf{Path Efficiency}~$\uparrow$ quantifies how direct successful trajectories are relative to the straight-line distance. Best in \textbf{bold}, second-best \underline{underlined}.
    }}
    \label{tab:navigation_quality}
    \Huge
    \renewcommand{\arraystretch}{1.3}
    \resizebox{\columnwidth}{!}{
    \begin{tabular}{
    l
    cccc c
    cccc c
    }
    \toprule
    \multirow{2}{*}{\textbf{Method}}
    & \multicolumn{5}{c}{\textbf{Median Steps to Goal} ($\downarrow$)}
    & \multicolumn{5}{c}{\textbf{Path Efficiency} ($\uparrow$)} \\
    \cmidrule(lr){2-6} \cmidrule(lr){7-11}
    & $L$=3 & $L$=5 & $L$=8 & $L$=12 & \textbf{Avg}
    & $L$=3 & $L$=5 & $L$=8 & $L$=12 & \textbf{Avg} \\
    \midrule
    LSTM
    & --- & 861 & \textbf{578} & 612 & 684 & --- & 0.666 & \textbf{0.905} & 0.852 & 0.807 \\
    Transformer
    & 636 & 644 & 607 & \underline{595} & 621 & 0.835 & 0.822 & 0.866 & \underline{0.881} & 0.851 \\
    Mamba
    & \underline{587} & \underline{592} & 605 & 640 & \underline{606} & \underline{0.895} & \underline{0.886} & 0.866 & 0.812 & \underline{0.865} \\
    \textbf{\ours{}}
    & \textbf{578} & \textbf{570} & \underline{588} & \textbf{567} & \textbf{576} & \textbf{0.915} & \textbf{0.918} & \underline{0.896} & \textbf{0.932} & \textbf{0.915} \\
    \bottomrule
    \end{tabular}
    }
    \vspace{-10pt}
    \end{table}

\textbf{Trajectory Smoothness:}   
Lastly, Table~\ref{tab:trajectory_smoothness} quantifies motion quality through SPARC and LDLJ metrics. \ours{} achieves the best SPARC ($-3.550$) and LDLJ ($-8.180$), outperforming Mamba by 1.0\% and 1.8\% respectively, with consistent best-in-class performance across all sequence lengths. Smoother trajectories reduce energy consumption by minimizing acceleration changes and improving actuator performance. \ours{} simultaneously achieves both fastest navigation and smoothest trajectories, indicating superior trajectory optimization rather than aggressive path-taking.

    \begin{table}[t!]
    \centering
    \caption{\small{
    Trajectory smoothness evaluation on successful episodes.
    \textbf{SPARC}~$\uparrow$ measures velocity smoothness~\cite{beck2018sparc}.
    \textbf{LDLJ}~$\uparrow$ quantifies jerk-based smoothness~\cite{gulde2018smoothness}.
    Best in \textbf{bold}, second-best \underline{underlined}.
    }}
    \label{tab:trajectory_smoothness}
    \Huge
    \renewcommand{\arraystretch}{1.5}
    \setlength{\tabcolsep}{3pt}
    \resizebox{\columnwidth}{!}{
    \begin{tabular}{
    p{4.2cm}
    cccc c
    cccc c
    }
    \toprule
    \multirow{2}{*}{\textbf{Method}}
    & \multicolumn{5}{c}{\textbf{SPARC} ($\uparrow$)}
    & \multicolumn{5}{c}{\textbf{LDLJ} ($\uparrow$)} \\
    \cmidrule(lr){2-6} \cmidrule(lr){7-11}
    & $L$=3 & $L$=5 & $L$=8 & $L$=12 & \textbf{Avg}
    & $L$=3 & $L$=5 & $L$=8 & $L$=12 & \textbf{Avg} \\
    \midrule
    LSTM
    & --- & -3.734 & \underline{-3.555} & -3.583 & -3.624 & --- & -8.970 & \textbf{-8.169} & -8.299 & -8.479 \\
    Transformer
    & -3.603 & -3.615 & -3.586 & \underline{-3.571} & -3.594 & -8.365 & -8.399 & -8.293 & \underline{-8.241} & \underline{-8.325} \\
    Mamba
    & \underline{-3.564} & \underline{-3.562} & -3.576 & -3.611 & \underline{-3.578} & \underline{-8.234} & \underline{-8.243} & -8.342 & -8.493 & -8.328 \\
    \textbf{\huge{\ours{}}}
    & \textbf{-3.553} & \textbf{-3.553} & \textbf{-3.552} & \textbf{-3.543} & \textbf{-3.550} & \textbf{-8.185} & \textbf{-8.159} & \underline{-8.228} & \textbf{-8.150} & \textbf{-8.180} \\
    \bottomrule
    \end{tabular}
    }
    \vspace{-15pt}
    \end{table}

\section{Real-World Deployment and Validation}
Beyond simulation, we demonstrate sim-to-real transfer by deploying the trained policy on a Unitree Go1 robot.

\subsection{Hardware and Deployment Setup}
The real-world deployment uses a Unitree Go1 quadruped equipped with an RSLidar-16 sensor. All sensor data streams are processed on an offboard laptop (Intel i7-12700H, NVIDIA RTX 3050-Ti) running a ROS2 package that assembles the concatenated policy observation at each control step. Specifically, the raw 3D point cloud from the RSLidar-16 is projected onto a horizontal plane to produce 360°  scans consistent with the 2D scan representation used during IsaacLab training. Robot localization is provided by a VICON motion capture system, supplying global position estimates that are transformed into the robot body frame at each step, while base linear velocity is estimated directly via the Unitree SDK. All observations are transmitted wirelessly to the offboard compute unit, where the trained policy runs at 5\,Hz, and issues velocity commands back to the Go1's onboard high-level controller through the Unitree SDK.

\begin{figure}[t!]
      \vspace{4pt}
      \centering    \includegraphics[width=0.97\columnwidth]{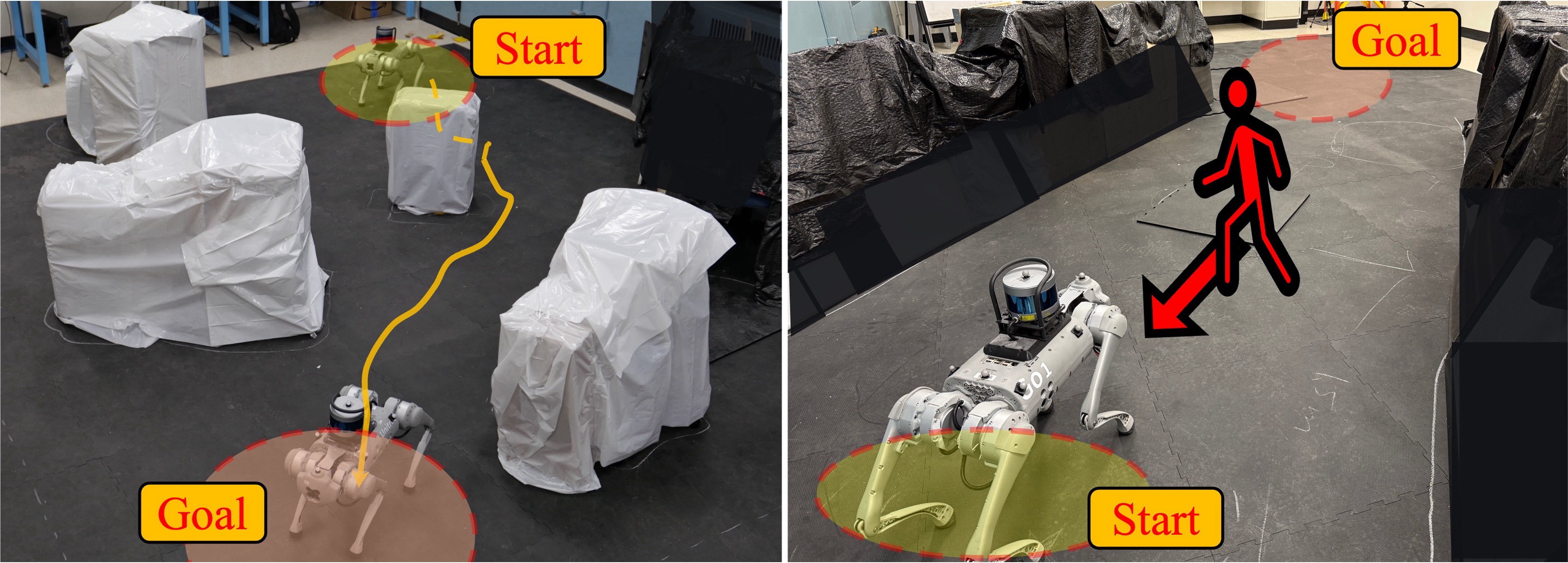}
      \caption{\small{Indoor real-world evaluation with the Unitree Go1 quadruped robot. \textbf{Left:} Static obstacle-rich environment. \textbf{Right:} Moving pedestrian approaching the robot.}
      \label{fig:realworld_scenarios}}
      \vspace{-16pt}
\end{figure}

Experiments are conducted in an indoor room of approximately 8\,$\times$\,6\,m; however, the effective VICON tracking volume covers approximately a central 6\,$\times$\,4.5\,m region, within which all trials are constrained. Goal positions are defined in the VICON world frame, with its origin at the center of the capture volume. Each trial commands the robot from a start position at $(-1.5,\,-2)$\,m to a goal at $(1.5,\,2)$\,m, spanning 5\,m of lateral travel within the effective tracking range. As illustrated in Figure~\ref{fig:realworld_scenarios}, we evaluate two obstacle scenarios: (1) \textit{static obstacles}, where the environment contains several fixed obstacles that block the robot's path, and (2) \textit{dynamic obstacle}, where a pedestrian walks from the origin toward the robot at approximately 0.15--0.2\,m/s within a constrained 3\,m-wide corridor, requiring reactive collision avoidance. For each scenario, we test under two LiDAR scan range configurations: 1.5\,m (short-range) and 3.0\,m (medium-range), yielding four evaluation conditions in total. A trial is considered successful if the robot reaches the goal within a radius of $0.4$\,m without direct collision or leaving the VICON tracking volume.

\subsection{Tuning Process}
Prior to evaluation, we clip the linear velocity to $\pm0.3$\,m/s and the angular velocity to $\pm0.6$\,rad/s to account for the constrained indoor environment, and we observe that clipping yields better behavioral consistency than action scaling.

\subsection{Evaluation Metrics}
We report Path Efficiency (Eq.~\ref{eq:path_efficiency}) for successful trials. We also evaluate overall navigation performance using Success-weighted Path Length (SPL) \cite{anderson2018evaluation}, which jointly captures task success and path efficiency. \vspace{-10pt}
\begin{equation}
\text{SPL}_i = S_i \cdot \frac{L^*_i}{\max(L^*_i, P_i)}, \quad \text{SPL} = \frac{1}{N}\sum_{i=1}^{N} \text{SPL}_i
\label{eq:spl_realworld}
\end{equation}
where $S_i \in \{0,1\}$ indicates success, $L^*_i$ is the obstacle-free shortest path length, and $N$ is the total number of trials.

\subsection{Quantitative Results}
As shown in Table~\ref{tab:realworld}, we evaluate each method across 20 trials per configuration. At short LiDAR range (1.5\,m), all methods achieve reasonable performance. \ours{} reaches 95\% success in both static and dynamic scenarios with best-in-class SPL, while vanilla Mamba achieves 85\% success and Transformer 55--60\%. However, extending the scan range to 3.0\,m reveals substantial performance degradation: Transformer and vanilla Mamba fail catastrophically in static environments (no success), and Mamba degrades severely in dynamic scenarios. In contrast, \ours{} maintains robust performance across both ranges, achieving the best success rate and SPL in all configurations.

\begin{table}[t]
\vspace{4pt}
\centering
\caption{\small{Real-world navigation on Unitree Go1 with RSLidar-16 ($L{=}12$, 20 trials each). Best in \textbf{bold}.}}
\label{tab:realworld}
\setlength{\tabcolsep}{3.5pt}
\renewcommand{\arraystretch}{0.6}
\small
\begin{tabular}{llccc}
\toprule
Scenario & Method & Success & Path Eff. & SPL \\
\midrule
\multirow{4}{*}{\shortstack[l]{Static Obstacles\\(LiDAR scan: 1.5\,m)}}
 & LSTM        & 6/20   & \textbf{0.919} & 0.276 \\
 & Transformer & 12/20  & 0.807          & 0.484 \\
 & Mamba       & 17/20  & 0.892          & 0.758 \\
 & \textbf{\ours{}}    & \textbf{19/20} & 0.879  & \textbf{0.835} \\
\midrule
\multirow{4}{*}{\shortstack[l]{Static Obstacles\\(LiDAR scan: 3.0\,m)}}
 & LSTM        & 6/20   & 0.748          & 0.224 \\
 & Transformer & 0/20   & ---            & ---   \\
 & Mamba       & 0/20   & ---            & ---   \\
 & \textbf{\ours{}}     & \textbf{18/20} & \textbf{0.895} & \textbf{0.805} \\
\midrule
\multirow{4}{*}{\shortstack[l]{Moving Pedestrian\\(LiDAR scan: 1.5\,m)}}
 & LSTM        & 4/20   & 0.933          & 0.187 \\
 & Transformer & 11/20  & 0.952          & 0.524 \\
 & Mamba       & 17/20  & 0.964          & 0.819 \\
 & \textbf{\ours{}}     & \textbf{19/20} & \textbf{0.967} & \textbf{0.919} \\
\midrule
\multirow{4}{*}{\shortstack[l]{Moving Pedestrian\\(LiDAR scan: 3.0\,m)}}
 & LSTM        & 10/20  & 0.926          & 0.463 \\
 & Transformer & 0/20   & ---            & ---   \\
 & Mamba       & 6/20   & 0.927          & 0.278 \\
 & \textbf{\ours{}}     & \textbf{17/20} & \textbf{0.932} & \textbf{0.792} \\
\bottomrule
\end{tabular}
\vspace{-17pt}
\end{table}

This performance gap may stem from the way each architecture handles the increased spatial complexity across different LiDAR ranges. At 3.0\,m, scans capture not only nearby obstacles but also distant walls, room boundaries, and the full spatial layout, requiring the network to reason over substantially more environmental context than under the 1.5\,m configuration, which primarily observes local obstacles. We hypothesize that integrating more distant and less immediately relevant measurements leads to progressive hidden-state corruption, manifesting as overly cautious obstacle avoidance without reliable goal recovery. Both methods detect obstacles effectively but fail to maintain goal-directed navigation when processing the full-room context. The proximity-gated stream in \ours{} continuously injects corrective LiDAR signals directly into each SSM layer, providing a dedicated pathway that filters and prioritizes obstacle-relevant information while mitigating error accumulation from distant readings. The superior SPL scores across all configurations indicate that \ours{} not only succeeds more reliably but also navigates more efficiently, demonstrating that the explicit dual-stream design effectively balances reactive safety with strategic navigation under challenging real-world sensing conditions.
\section{Conclusions}
We introduced \ours{}, a dual-stream state-space architecture for LiDAR-based obstacle avoidance that partitions Mamba's temporal encoding into specialized proximity and goal streams. Through controlled comparisons, \ours{} demonstrates advantages: the lowest timeout rate (0.4\%), the fastest navigation (576 median steps), the highest path optimality (0.915 path efficiency), and 1.5× faster training convergence across all sequence lengths. Beyond simulation, \ours{} successfully transfers across platforms—from quadrupedal Go2 in IsaacLab to wheeled Husky in Gazebo, and to real-world Go1 deployment with various layout setups. Real-world evaluation shows that \ours{} maintains robust performance across LiDAR scan ranges, whereas vanilla Mamba and Transformer exhibit severe performance degradation, which we hypothesize is associated with hidden-state corruption. The dual-stream design introduces a principled inductive bias for multi-rate temporal tasks: complementary specialization through asymmetric input routing yields measurable optimization and performance advantages over single-stream encoders.

\bibliographystyle{IEEEtran}
\bibliography{References}

\end{document}